%% file: conference_101719.tex
\documentclass[conference]{IEEEtran}
\usepackage{cite}
\usepackage{amsmath,amssymb,amsfonts}
\usepackage{url}
\usepackage[ruled,vlined,linesnumbered]{algorithm2e}
\usepackage{graphicx}
\usepackage{textcomp}
\usepackage{xcolor}
\usepackage{booktabs}
\usepackage{multirow}
\usepackage{colortbl}
\usepackage{array}
\usepackage{amsthm}

\usepackage[caption=false,font=footnotesize]{subfig}
\usepackage{xspace}
\newcommand{\methodname}{\textsc{VisPath}\xspace}
\def\BibTeX{{\rm B\kern-.05em{\sc i\kern-.025em b}\kern-.08em
    T\kern-.1667em\lower.7ex\hbox{E}\kern-.125emX}}
\begin{document}

\newcommand{\zp}[1]{\textcolor{magenta}{(ZhP: #1)}}
\makeatletter
\newcommand{\linebreakand}{%
  \end{@IEEEauthorhalign}
  \hfill\mbox{}\par
  \mbox{}\hspace*{-1.0cm}\hfill\begin{@IEEEauthorhalign}
}
\makeatother

\title{\methodname{}: Visual-Intent-Guided Path Reasoning for Multimodal Knowledge Graph Question Answering
}

\author{\IEEEauthorblockN{1\textsuperscript{st} Jinke Wu\textsuperscript{*}}
\IEEEauthorblockA{
\textit{Sichuan University}\\
Chengdu, China \\
w\_rmsl@stu.scu.edu.cn}
\and
\IEEEauthorblockN{2\textsuperscript{nd} Zhengpin Li\textsuperscript{*}}
\IEEEauthorblockA{
\textit{Peking University}\\
Beijing, China \\
zpli@pku.edu.cn}
\and
\IEEEauthorblockN{3\textsuperscript{rd} Mengzhe Jia}
\IEEEauthorblockA{
\textit{Fudan University}\\
Shanghai, China \\
mzjia23@m.fudan.edu.cn}
\linebreakand
\and
\IEEEauthorblockN{4\textsuperscript{th} Yang Li}
\IEEEauthorblockA{\textit{Tencent} \\
Shenzhen, China \\
thomasyngli@tencent.com}
\and
\IEEEauthorblockN{5\textsuperscript{th} Wentao Zhang}
\IEEEauthorblockA{
\textit{Peking University}\\
Beijing, China \\
wentao.zhang@pku.edu.cn}
\thanks{\textsuperscript{*}Jinke Wu and Zhengpin Li contributed equally to this work.}
}

\maketitle
\begingroup
\renewcommand\thefootnote{*}
\footnotetext{Jinke Wu and Zhengpin Li contributed equally to this work.}
\endgroup

\input{sec/0_abstract}


\input{sec/1_intro}
\input{sec/2_related_work}
\input{sec/3_preliminaries}

\input{sec/4_method}

\input{sec/5_experiments}

\input{sec/6_conclusion}

\clearpage
\bibliographystyle{IEEEtran}
\bibliography{refs.bib}

\end{document}

%% file: sec/0_abstract.tex
\begin{abstract}
Knowledge graph question answering (KGQA) enables models to answer natural-language questions through structured graph reasoning and has achieved substantial progress across many benchmarks and applications. Recently, multimodal KGQA (MM-KGQA) has attracted increasing attention because many questions require jointly using multimodal inputs and KG evidence. However, existing MM-KGQA methods typically use multimodal information only for starting entity grounding or evidence retrieval, after which multi-hop reasoning degenerates into text-only graph search. As a result, they cannot exploit multimodal cues that become important at intermediate hops. To address this limitation, we propose \methodname{}, a visual-intent-guided path reasoning framework for MM-KGQA. \methodname{} first identifies a reliable starting entity by combining multimodal grounding with graph-structural cues. It then performs intent-guided path discovery by recomputing hop-specific multimodal intent from the input, question, and current partial paths, so that each expansion is guided by the current reasoning state. The discovered paths are further refined through reasoning-chain pruning, which evaluates candidate paths as complete evidence chains based on their consistency with the question, reasoning sketch, and hop-specific intent. Finally, \methodname{} checks whether the selected evidence is sufficient for answer generation. We further construct \methodname{}-Bench, a benchmark for evaluating multimodal multi-hop reasoning over KGs, covering questions that require two to four hops over KG paths. Extensive experiments on \methodname{}-Bench and three additional multimodal QA benchmarks show that \methodname{} consistently outperforms strong KGQA, vision--language, and retrieval-augmented baselines. Notably, with GPT-4o as the backbone, \methodname{} surpasses GPT-5.4 on \methodname{}-Bench, achieving a 10.6\% relative improvement in average accuracy and a 13.1\% improvement at 2-hop reasoning.
\end{abstract}

%% file: sec/1_intro.tex
\section{Introduction}

Knowledge graphs (KGs) have become a widely adopted form of structured knowledge representation~\cite{hogan2021knowledge,mensink2023encyclopedic}. By organizing real-world knowledge into entities, relations, and factual triples, KGs provide an explicit and interpretable foundation for knowledge access and reasoning. Due to their structured nature, KGs have been widely used in search engines, recommender systems, question answering, and knowledge-intensive reasoning tasks~\cite{pan2024unifying,li2024spectral,zhang2026mmhops, tran2025reasonvqa, li2024beyond}. Compared with unstructured text corpora, KGs make relational dependencies more explicit and allow the reasoning process to be traced through supporting facts~\cite{luo2024rog,li2025fairness}.  

One important application of KGs is knowledge graph question answering (KGQA). KGQA aims to answer natural-language questions by locating relevant entities, traversing relations, and reasoning over graph facts. For example, given the question ``Where was the architect of the Government Museum, Chandigarh born?'', a KGQA system needs to first locate the entity \textit{Government Museum, Chandigarh}, follow the relation \textit{architect} to \textit{Le Corbusier}, and then follow \textit{place of birth} to infer the answer \textit{La Chaux-de-Fonds}. This example illustrates the key strength of KGQA: the answer is not generated only from textual patterns, but is derived through an explicit reasoning path over structured facts. Existing KGQA methods have achieved substantial progress, ranging from semantic parsing~\cite{berant2013semantic,yih2015semantic} and neural graph reasoning~\cite{sun2018graftnet,he2021improving} to recent LLM-based graph exploration methods~\cite{sun2024tog,chen2024plangraph,tan2025paths}. With these advances, KGQA methods have achieved state-of-the-art performance across many benchmarks and application scenarios, demonstrating the effectiveness of structured graph reasoning for knowledge-intensive question answering.

Despite the progress of KGQA, most existing methods are designed for unimodal settings~\cite{sun2024tog,luo2024rog,tan2025paths}. They typically assume that both the question and the knowledge graph are textual or symbolic, and that the starting entity can be identified from the question itself. However, many real-world questions are naturally multimodal, where the entity to be queried may only be specified by an image rather than by its textual name. In such cases, a model must first ground the visual content to the correct graph entity and then reason over the KG to obtain the answer. This makes standard KGQA methods insufficient, since they do not explicitly model the interaction between visual grounding and graph reasoning.

Recently, several studies have begun to explore multimodal knowledge graph question answering (MM-KGQA). In this task, a model is required to answer a question by jointly using visual inputs and structured KG evidence. Existing efforts show that multimodal evidence is important for connecting visual content with encyclopedic knowledge and for answering questions beyond what can be inferred from the image alone. Some methods construct multimodal KGs from documents or retrieve entity- and relation-level evidence for visual question answering~\cite{yuan2025mkgrag}. Meanwhile, several benchmarks have been proposed to evaluate multi-hop visual questions that require external knowledge beyond the image~\cite{tran2025reasonvqa,zhang2026mmhops,mensink2023encyclopedic}.

However, existing MM-KGQA methods still have an essential limitation. Visual information is often used only at the beginning of the reasoning process, such as for starting entity grounding or evidence retrieval. Once the starting entity or relevant evidence is obtained, the subsequent reasoning process usually degenerates into text-only graph search. This decoupled paradigm makes later reasoning heavily dependent on the correctness of the initial grounding. More importantly, it prevents the model from using visual cues that may become relevant at intermediate hops, even though such cues can be crucial for selecting the correct relation or path.

To address this limitation, we propose \methodname{}, a visual-intent-guided path reasoning framework for MM-KGQA. Instead of treating multimodal information as a one-time signal for starting entity grounding, \methodname{} keeps the interaction among the multimodal input, the question, and the multimodal KG~(MMKG) throughout the reasoning process. First, for starting entity identification, \methodname{} uses the multimodal referring expression in the given question to guide object recognition from the multimodal input, and combines the recognized object with graph-structural cues from the graph to select an entity that is both multimodally consistent and reasoning-supportive. Next, for path discovery, it recomputes a hop-specific intent from the multimodal input, the question, and the current partial paths, so that each expansion step is guided by the current multimodal reasoning state rather than by the question alone. Then, for reasoning-chain pruning, it evaluates candidate paths as complete evidence chains by considering their consistency with the question, the global reasoning sketch, and the multimodal intent of the current hop. Finally, before producing the answer, \methodname{} summarizes the selected graph paths under the multimodal context and checks whether the resulting multimodal evidence is sufficient to support answer generation. In this way, \methodname{} enables multimodal information to participate not only in entity grounding, but also in relation selection, path pruning, and answer verification, leading to reasoning paths that are both KG-faithful and multimodally grounded across the entire reasoning process.

We further introduce \methodname{}-Bench, a new dataset designed to evaluate multimodal multi-hop reasoning over KGs. Existing MM-KGQA datasets mainly focus on shallow reasoning~\cite{mensink2023encyclopedic,tran2025reasonvqa}, making it difficult to assess whether a model can maintain multimodal grounding across deeper graph paths. To fill this gap, \methodname{}-Bench contains questions that require two to four hops of reasoning over KG paths. Each question is constructed so that the starting entity cannot be directly read from the question text, but must be grounded from the multimodal input, while the final answer must be derived through structured graph reasoning. In this way, \methodname{}-Bench provides a more challenging and systematic resource for evaluating whether models can jointly perform multimodal grounding, multi-hop path reasoning, and evidence-faithful answer generation under realistic and complex settings.

Our contributions are summarized as follows. 
\begin{itemize}
    \item We propose \methodname{}, a visual-intent-guided path reasoning method that keeps visual evidence involved in hop-by-hop KG reasoning.
    \item We construct \methodname{}-Bench, a benchmark for evaluating image-grounded multi-hop reasoning over MMKGs.
    \item Extensive experiments show that \methodname{}, using GPT-4o as the LLM backbone, consistently outperforms strong baselines, surpassing GPT-5.4~\cite{openai2026gpt54} on \methodname{}-Bench with a 10.6\% relative improvement in average accuracy and a 13.1\% relative improvement at 2-hop reasoning.
\end{itemize}

%% file: sec/2_related_work.tex
\section{Related Work}

\subsection{Multimodal Knowledge Graphs}
\label{sec:related_mmkg}

Multimodal knowledge graphs (MMKGs) extend conventional knowledge graphs (KGs) by associating entities, relations, or triples with evidence from non-textual modalities, such as images, audio, and video~\cite{liang2024MMKGsurvey,zhu2022multi}. Existing MMKG resources differ mainly in the type of multimodal evidence they attach and the granularity at which such evidence is organized. At the entity level, IKRL~\cite{xie2017ikrl} pioneers image-enhanced entity representations by encoding entity images into the embedding space. MMKG~\cite{liu2019mmkg} aligns FB15K~\cite{bordes2013translating} with DBpedia~\cite{auer2007dbpedia} and YAGO~\cite{suchanek2007yago}, and enriches each entity with images and numerical literals. Richpedia~\cite{wang2020richpedia} links encyclopedic entities to images crawled at scale, while VisualSem~\cite{alberts2021visualsem} grounds a multilingual entity inventory built on BabelNet~\cite{navigli2012babelnet}. Later resources move beyond whole-entity images toward finer and richer grounding. AspectMMKG~\cite{zhang2023aspectmmkg} attaches aspect-specific images to capture different facets of an entity. VISTA~\cite{lee2023vista} grounds not only entities but also relational triples with visual evidence. TIVA-KG~\cite{wang2023tiva} extends triple-level grounding across text, image, video, and audio. MMEKG~\cite{ma2022mmekg} further builds an event-centric MMKG that connects textual and visual events. Another line of work organizes multimodal knowledge for unified representation or downstream use. Examples include UKnow~\cite{gong2024uknow} for unified entity- and concept-level cross-modal knowledge, M\textsuperscript{2}ConceptBase~\cite{zha2024m2conceptbase} for fine-grained concept-image alignment, and VAT-KG~\cite{park2025vat} for concept-centric multimodal knowledge.

\subsection{Retrieval-Augmented Generation}
\label{sec:related_rag}

Retrieval-augmented generation (RAG) grounds large language model~(LLM) outputs in retrieved external evidence to reduce hallucination on knowledge-intensive tasks~\cite{lewis2020rag}. Later studies make retrieval more adaptive and agentic. Self-RAG~\cite{asai2024selfrag} retrieves evidence on demand and critiques its own generations through reflection tokens, while Search-o1~\cite{li2025searcho1} integrates an agentic search workflow into long-form reasoning. In the multimodal setting, OmniSearch~\cite{li2025OmniSearch} plans sub-question chains with retrieval actions for dynamically changing visual questions, and MMSearch-R1~\cite{wu2025mmsearch} trains a multimodal model with reinforcement learning to perform on-demand, multi-turn image and text search. Another line of work incorporates graph structures into retrieval. GraphRAG~\cite{edge2024graphrag} builds an LLM-derived entity graph over a text corpus and answers global questions through hierarchical community summaries. LightRAG~\cite{guo2025lightrag} couples graph-structured text indexing with a dual-level retrieval scheme, and G-Retriever~\cite{he2024gretriever} retrieves question-relevant subgraphs from textual graphs for answering.

\noindent\textbf{Limitation.} Traditional RAG retrieves free-form textual evidence, making it difficult to organize and verify the relational dependencies required by multi-hop questions. Graph-based RAG introduces graph structures into retrieval, but they are mainly constructed for conventional KGs and not designed for MMKGs, where reasoning requires grounding visual evidence.

\subsection{Knowledge Graph Question Answering}
\label{sec:related_kgqa}

\noindent\textbf{Traditional KGQA.} Knowledge graph question answering (KGQA) aims to answer natural-language questions by reasoning over a structured knowledge graph. Before the rise of LLMs, KGQA methods mainly followed two paradigms: semantic parsing and information-retrieval-based reasoning. Semantic parsing methods translate a question into an executable logical form or query graph. For example, SEMPRE~\cite{berant2013semantic} learns logical-form mapping from question--answer pairs over Freebase~\cite{bollacker2008freebase}, while STAGG~\cite{yih2015semantic} formulates KGQA as staged query-graph generation. Later methods, such as QGG~\cite{lan2020query} and RNG-KBQA~\cite{ye2022rng}, further improve query-graph construction and candidate ranking. Another line of work retrieves or propagates information over a question-specific subgraph. KV-MemNN~\cite{miller2016kvmem} reads facts from a key--value memory. GraftNet~\cite{sun2018graftnet} performs graph convolution over a fused KB-and-text subgraph, and PullNet~\cite{sun2019pullnet} further retrieves such subgraphs iteratively for multi-hop questions. EmbedKGQA~\cite{saxena2020embedkgqa} matches questions with pretrained KG embeddings to reach distant answers. NSM~\cite{he2021improving} and TransferNet~\cite{shi2021transfernet} perform multi-step relation reasoning with intermediate supervision. 

\noindent\textbf{LLM-based KGQA.} LLMs have reshaped KGQA by enabling models to reason over KG facts in natural-language form~\cite{pan2024unifying}. Early LLM-based methods retrieve question-relevant triples and provide them to the model as textual context~\cite{baek2023knowledge}. To better exploit graph structure, later studies treat the LLM as an agent that iteratively explores the KG. Think-on-Graph~\cite{sun2024tog} performs beam search over relations and entities from the starting entity. Reasoning-on-Graph~\cite{luo2024rog} first generates relation paths as plans and then retrieves supporting facts. Chain-of-Knowledge~\cite{li2024chain} interleaves structured and unstructured retrieval during chain-of-thought reasoning. Think-on-Graph 2.0~\cite{ma2025tog2} further couples graph traversal with document retrieval. More recent efforts emphasize adaptive planning and path-level reasoning. Plan-on-Graph~\cite{chen2024plangraph} decomposes a question into sub-objectives and self-corrects erroneous exploration. Paths-over-Graph~\cite{tan2025paths} constructs multi-hop reasoning paths and prunes them using graph structure. A few recent studies introduce multimodal evidence into multimodal knowledge graph question answering~(MM-KGQA). mKG-RAG~\cite{yuan2025mkgrag} constructs MMKGs from documents and retrieves entity- and relation-level evidence for visual question answering. ReasonVQA~\cite{tran2025reasonvqa} and MMhops-R1~\cite{zhang2026mmhops} further build large-scale benchmarks whose questions require multi-hop reasoning grounded in encyclopedic knowledge beyond the image itself.

\noindent\textbf{Limitation.}
Despite this progress, existing MM-KGQA methods typically use visual information only for starting entity grounding. After that, reasoning still largely degenerates into text-only graph search, making it difficult to exploit visual cues that become relevant at intermediate hops.

%% file: sec/3_preliminaries.tex
\section{Preliminaries}
\label{sec:prelim}

\subsection{Multimodal Knowledge Graph}
\label{sec:prelim-kg}

We denote a multimodal knowledge graph (MMKG) as
$\mathcal{G}=(\mathcal{E},\mathcal{R},\mathcal{F},\mathcal{X},\mathcal{V})$,
where $\mathcal{E}$, $\mathcal{R}$, and $\mathcal{F}$ are the sets of entities, relations, and facts, respectively. $\mathcal{X}$ denotes the set of visual resources, and $\mathcal{V}$ maps each entity to its associated visual resources. In this paper, we focus on the visual modality following existing studies on MMKGs. Each fact in $\mathcal{F}$ is represented as a triple $(e_{\mathrm{subj}},r,e_{\mathrm{obj}})$, where $e_{\mathrm{subj}},e_{\mathrm{obj}}\in\mathcal{E}$ are the subject and object, respectively, and $r\in\mathcal{R}$. A reasoning path of length $d$ in an MMKG is denoted as
$p=(e_0,r_1,e_1,\ldots,r_d,e_d)$,
where $(e_{i-1},r_i,e_i)\in\mathcal{F}$ for every $i=1,\ldots,d$. Here, $e_0$ is the starting entity of the path, and $e_d$ is the endpoint entity. For example,
$\textit{Government Museum, Chandigarh} \xrightarrow{\textit{architect}} \textit{Le Corbusier} \xrightarrow{\textit{place of birth}} \textit{La Chaux-de-Fonds}$
is a length-2 path, where $\textit{Government Museum, Chandigarh}$ is the head entity and $\textit{La Chaux-de-Fonds}$ is the endpoint entity.

\subsection{Problem Formulation}
\label{sec:prelim-problem}

In this paper, we study multimodal knowledge graph question answering (MM-KGQA). Given a multimodal query $(I,Q,\mathcal{G})$, where $I$ is an input image, $Q$ is a natural-language question related to $I$, and $\mathcal{G}$ is an MMKG, the goal is to return the correct answer $A$ by jointly grounding the visual content in $I$ and reasoning over $\mathcal{G}$. The answer should be supported by a reasoning path in $\mathcal{G}$ that starts from the image-related starting entity and reaches evidence relevant to $Q$. For example, given an image $I$ of \textit{Government Museum, Chandigarh} and the question $Q$ ``Where was the architect of the building shown in the image born?'', the model is expected to identify \textit{Government Museum, Chandigarh} as the starting entity, follow the path
$\textit{Government Museum, Chandigarh} \xrightarrow{\textit{architect}} \textit{Le Corbusier} \xrightarrow{\textit{place of birth}} \textit{La Chaux-de-Fonds}$,
and return the answer $A=\textit{La Chaux-de-Fonds}$.

%% file: sec/4_method.tex
\input{fig/framework}
\section{Method}
\label{sec:method}
Existing MM-KGQA methods typically use multimodal information, such as visual cues from the image $I$, only for starting entity grounding. Once the starting entity $e_0$ is identified, graph reasoning is reduced to triple-only path search over $(e_0, Q, \mathcal{G})$. The subsequent process expands candidate relations, reaches intermediate entities, and selects reasoning paths solely within the knowledge graph, without further access to $I$. Such a decoupled process makes later reasoning heavily dependent on the correctness of the initial grounding. It also prevents the model from using visual cues that may become important at intermediate hops.

To address this limitation, we propose a novel MM-KGQA method named \methodname{}, which answers a multimodal query $(I, Q, \mathcal{G})$ by keeping visual evidence involved throughout the reasoning process. Specifically, \methodname{} first performs \emph{visual- and graph-aware starting entity identification} to locate the image-related starting entity, providing a reliable graph entry point for subsequent multi-hop reasoning. Starting from this entity,  it then conducts \emph{intent-guided reasoning path discovery}, which progressively explores the MMKG by guiding each path expansion toward image-relevant paths using visual intent.  The discovered paths are further processed by \emph{reasoning-chain pruning}, which evaluates candidates as complete reasoning chains rather than isolated edges and retains the most globally coherent paths.  Finally, \methodname{} performs \emph{visual-grounded question answering}, where the selected paths are summarized with respect to the visual content, checked for answerability, and then used to either produce the final answer or continue the search when the current evidence is insufficient. In this way, \methodname{} reduces the dependence on perfect initial grounding, keeps graph reasoning continuously aligned with visual evidence, and produces answers that are both path-faithful and image-grounded. We illustrate \methodname{} in Figure~\ref{fig:framework} and provide the whole reasoning process in Algorithm~\ref{alg:vispath}.

% --- 4.1 --------------------------------------------------------------
\subsection{Visual- and Graph-aware Starting Entity Identification}
\label{sec:method-grounding}

Existing MM-KGQA methods usually identify the starting entity by matching the visual content of the image with entity names in the MMKG. Relying only on such visual-name matching can be unreliable, since the object recognized from the image may be described with a different name, alias, or level of granularity from the entity stored in the graph. Moreover, even when an entity has the highest visual-name similarity, it may not provide relations that are useful for answering $Q$. Selecting such an entity as the starting point can mislead the subsequent graph reasoning process. To address this issue, \methodname{} identifies the starting entity $e_S$ for the query $(I,Q,\mathcal{G})$ by jointly considering visual evidence and graph-structural cues. In this way, the selected entity is not only visually consistent with the image, but also structurally suitable as the entry point for multi-hop reasoning.

\noindent\textbf{Visual-aware candidate ranking.}
We first prompt a text language model $\texttt{LLM}(\cdot)$ to extract the expression in $Q$ that refers to the visual object in $I$, together with its bare head noun, denoted as $\mathrm{HEAD}$. We use only $\mathrm{HEAD}$ for visual recognition, so that the vision model focuses on the object depicted in the image rather than being biased by other reasoning clues in $Q$. Given the image $I$ and the extracted $\mathrm{HEAD}$, we use a vision--language model $\texttt{VLM}(\cdot)$ to recognize the depicted object and produce a textual object name:
\begin{equation}
\label{eq:visual-recog}
o = \texttt{VLM}\left(I, \mathrm{HEAD}\right),
\end{equation}
where $o$ denotes the object name predicted from the image. Since $o$ is generated by visual recognition, it may not exactly match the entity names stored in the MMKG. We therefore encode $o$ with a sentence encoder $\texttt{SenBERT}(\cdot)$~\cite{reimers2019sentencebert} and retrieve similar entity names from $\mathcal{G}$ using a pre-built entity-name embedding index. We then provide the retrieved entity names, the predicted object name $o$, and the question $Q$ to $\texttt{LLM}(\cdot)$. The language model filters and ranks the retrieved entities by considering both their semantic similarity to $o$ and their compatibility with the image-referring expression in $Q$. The ranked entities returned by $\texttt{LLM}(\cdot)$ form the initial candidate set, denoted as $\{e_{Q,1},\dots,e_{Q,K}\}$, where earlier entities are considered more likely to be the starting entity for answering $Q$. Rather than directly accepting the top-ranked entity, we keep the entire ranked candidate set for the subsequent validation step. This preserves alternative visually plausible entities and allows revising the initial ranking when the top candidate is not suitable for downstream reasoning.

\noindent\textbf{Graph-aware validation.}
Visual similarity can identify entities that are visually or lexically close to the object in the image, but it does not indicate whether these entities are suitable starting points for answering $Q$. This distinction is important in MMKG reasoning. A visually plausible candidate may be disconnected from the required answer path, or it may only expose outgoing relations that are irrelevant to the question. Therefore, we further validate each candidate by checking whether its outgoing relations are compatible with the reasoning need expressed in $Q$. For each candidate $e_{Q,k}$, we compute a graph-aware alignment score:
\begin{equation}
\label{eq:edge-align}
s_{Q,k} = \max_{r \in \mathcal{R}_{e_{Q,k}}} \cos\big(\texttt{SenBERT}(r), \texttt{SenBERT}(Q)\big),
\end{equation}
where $\mathcal{R}_{e_{Q,k}}$ denotes the set of outgoing relations of $e_{Q,k}$. This score measures whether the candidate provides at least one outgoing relation that can plausibly initiate the answer path. We use this score to correct the visual linking result. Denote $e_{Q,1}$ as the top-ranked candidate from visual-aware candidate linking. Let  $e_{Q,k^\star}$ be the candidate with the highest graph-aware alignment score, where $k^\star=\arg\max_{2\leq k\leq K} s_{Q,k}$. We replace $e_{Q,1}$ with $e_{Q,k^\star}$ only when the latter is clearly more suitable for graph reasoning:
\begin{equation}
\label{eq:graph-aware-correction}
s_{Q,k^\star}-s_{Q,1}>\delta,
\quad \delta>0.
\end{equation}
Otherwise, we keep $e_{Q,1}$ as the starting entity. After this correction, we obtain an updated ranked candidate list $\{ \tilde{e}_{Q,1},\dots,\tilde{e}_{Q,K}\}$ and take its first element as the starting entity, i.e., $e_S=\tilde{e}_{Q,1}$. This conservative correction avoids changing reliable visual links unnecessarily, while still allowing the model to recover when the visually top-ranked entity is not structurally useful for subsequent reasoning.

% --- 4.2 --------------------------------------------------------------

\subsection{Intent-Guided Reasoning Path Discovery}
\label{sec:method-discovery}
After identifying the starting entity $e_S$, existing KGQA methods usually rely only on the question $Q$ to determine which relations should be followed in subsequent hops. As a result, path search is conducted purely over the graph, and the image is no longer involved in the reasoning process. This question-only strategy is insufficient for MMKG reasoning, because the question may not explicitly specify every relation needed in the answer path. In such cases, some intermediate reasoning clues must still be inferred from the image. For example, consider the question ``\textit{Which league does the team shown on the player's jersey belong to?}'' Even if the player is correctly grounded as the starting entity, the graph may contain many outgoing relations, such as \textit{place of birth}, \textit{position}, \textit{national team}, and multiple \textit{club} relations. The question does not explicitly indicate which team should be followed. Instead, the relevant team must be identified from the jersey logo or other visual details in the image. Therefore, visual evidence remains necessary after grounding: it guides the first hop toward the visually indicated \textit{club} relation, and the next hop can then follow the \textit{league} relation from that club.

To this end, we propose an intent-guided reasoning path discovery module that keeps the image involved after the starting entity has been identified. Starting from $e_S$, the module progressively discovers reasoning paths over the MMKG. At each hop, it infers the reasoning intent for the next step by jointly considering the image $I$, the question $Q$, and the current partial paths. This intent is then used to select the most relevant outgoing edges from the current entities. The selected edges extend the current paths and produce a new set of candidate reasoning paths for the next hop. In this way, \methodname{} can use visual evidence to decide which relation should be followed at each reasoning state, avoid blindly expanding irrelevant graph branches, and discover reasoning paths that are both graph-valid and image-relevant.

Formally, let $\mathcal{P}^{(i-1)}$ denote the set of partial reasoning paths retained after hop $i-1$, and let $\mathcal{E}^{(i-1)}$ denote their endpoint entities. We initialize the search with $\mathcal{P}^{(0)}=\{(e_S)\}$ and $\mathcal{E}^{(0)}=\{e_S\}$. At each hop $i$, we first use $\texttt{VLM}(\cdot)$ to generate a hop-specific reasoning intent:
\begin{equation}
\label{eq:hop-intent}
q_{Q}^{(i)} = \texttt{VLM}\left(I, Q, \mathcal{P}^{(i-1)}\right),
\end{equation}
where $q_{Q}^{(i)}$ is a short textual description of what the next reasoning step should seek. Since this intent is conditioned on the image, the question, and the current partial paths, it can adapt to different reasoning stages. In the jersey example, the first hop may focus on the club indicated by the player's jersey, while the next hop focuses on the league associated with that club.

Given the hop-specific intent $q_{Q}^{(i)}$, we compute an intent-matching score for each outgoing edge from the current endpoint entities. For an entity $e\in\mathcal{E}^{(i-1)}$ and one of its outgoing edges $(e,r,e_t)$, we define:
\begin{equation}
\label{eq:edge-score}
c_{r,q_i} =\cos\left(
\texttt{SenBERT}\left(r \mathbin{\Vert} e_t\right),
\texttt{SenBERT}\left(q_{Q}^{(i)}\right)
\right),
\end{equation}
where $\mathbin{\Vert}$ denotes textual concatenation, and $e_t$ is the endpoint entity reached by relation $r$. We include the endpoint entity together with the relation because the relation name alone may be too coarse to distinguish visually relevant edges from irrelevant ones. For each entity $e\in\mathcal{E}^{(i-1)}$, we keep only the top-$M_1$ outgoing edges with the highest intent-matching scores. The collected edge set is denoted as $\mathcal{R}^{(i-1)}$. This operation limits the branching factor at each hop and prevents high-degree entities from introducing excessive irrelevant candidates.

Finally, we construct new candidate paths by extending each retained partial path with the selected outgoing edges from its endpoint entity:
\begin{equation}
\label{eq:path-extension}
{\mathcal{P}}^{(i)}_{\mathrm{cand}} =\bigcup_{p\in\mathcal{P}^{(i-1)}}
\left\{
p\oplus(\operatorname{end}(p),r,e_t)
\mid
r\in\mathcal{R}^{(i-1)}
\right\}.
\end{equation}
Here, $\operatorname{end}(p)$ denotes the endpoint entity of path $p$, and $p\oplus(\operatorname{end}(p),r,e_t)$ appends relation $r$ and endpoint entity $e_t$ to $p$. Since at most $M_1$ outgoing edges are retained for each endpoint entity, the number of generated candidate paths is bounded by
\begin{equation}
\label{eq:path-bound}
\left|\mathcal{P}_{\mathrm{cand}}^{(i)}\right|
\leq
M_1\left|\mathcal{P}^{(i-1)}\right|.
\end{equation}
This bounded construction allows \methodname{} to explore visually relevant paths hop by hop while controlling the growth of the candidate space.

% --- 4.3 --------------------------------------------------------------
\subsection{Reasoning-Chain Pruning}
\label{sec:method-selection}

The intent-guided discovery module generates a candidate path set $\mathcal{P}_{\mathrm{cand}}^{(i)}$ at hop $i$. Although each expanded edge is selected according to the hop-specific intent, such local matching does not guarantee that the whole path is suitable for answering the question. A path may contain edges that are individually plausible, but their composition may deviate from the expected reasoning direction. In other words, locally optimal expansions at different hops may not form a globally coherent reasoning chain. Therefore, we further prune and select candidate paths at the reasoning-chain level, so that the retained paths are consistent with the question $Q$, the visual evidence from $I$, and the hop-specific intent $q_Q^{(i)}$ as a whole.

To provide a global reference for chain-level selection, we first use $\texttt{LLM}(\cdot)$ to analyze the question $Q$ into a reasoning sketch $\mathcal{S}_Q$ and a predicted reasoning depth $\widehat{D}$. We use $\widehat{D}$ only when it is smaller than the pre-defined maximum search depth $D_{\max}$; otherwise, the search is bounded by $D_{\max}$. The sketch describes the coarse relation-level reasoning direction, while $\widehat{D}$ estimates the number of hops required to reach the answer. For example, for the question ``Where was the architect of the building shown in the image born?'', the question can be summarized into a two-hop sketch, `architect $\rightarrow$ place of birth'', with $\widehat{D}=2$. This sketch does not determine the exact entities in the path, but it provides a global semantic guide for judging whether a candidate path follows the expected reasoning chain.

To evaluate whether a candidate path is globally coherent, we compare the whole path with the multimodal reasoning objective. Specifically, we first convert each candidate path into a textual relation chain, and then measure its semantic consistency with the question, the reasoning sketch, and the hop-specific visual intent. For each path $p\in\mathcal{P}_{\mathrm{cand}}^{(i)}$, the chain-level score is defined as
\begin{equation}\footnotesize
\label{eq:path-rerank}
\sigma_{p,i}=\cos\left(
\texttt{SenBERT}\left(\texttt{Verbalize}(p)\right),
\texttt{SenBERT}\left(Q \mathbin{\Vert} \mathcal{S}_Q \mathbin{\Vert} q_Q^{(i)}\right)
\right),
\end{equation}
where $\texttt{Verbalize}(p)$ transforms the structured path $p$ into a natural-language relation chain. Here, $Q$ specifies the answer objective, $\mathcal{S}_Q$ provides the global reasoning sketch, and $q_Q^{(i)}$ provides the image-derived intent at hop $i$. Thus, $\sigma_{p,i}$ measures whether the complete path is compatible with the multimodal reasoning goal, rather than only checking whether its latest edge matches the current hop intent. We then select the top-$M_2$ paths according to $\sigma_{p,i}$ and denote the selected path set as $\mathcal{P}_{\mathrm{chain}}^{(i)}$. This step removes paths whose complete relation chains are weakly aligned with the expected reasoning direction, even if some of their individual edges were selected by the hop-level intent.

After semantic scoring, we further employ a language model to examine the remaining paths as complete reasoning chains and select those that are most likely to provide valid evidence for answering the question. If the language model returns an empty set, we fall back to the highest-scoring paths in $\mathcal{P}_{\mathrm{chain}}^{(i)}$. We then retain at most $M_3$ paths according to their chain-level scores and denote the retained path set as $\mathcal{P}^{(i)}$. By moving from locally intent-matched edge expansion to globally coherent chain selection, \methodname{} reduces the risk of accumulating plausible but inconsistent reasoning steps, while keeping the search focused on paths that are both visually grounded and semantically aligned with the question.

% --- 4.4 --------------------------------------------------------------
\subsection{Visual-Grounded Question Answering}
\label{sec:method-answering}

After reasoning-chain pruning and selection, \methodname{} obtains a retained path set $\mathcal{P}^{(i)}$ at hop $i$. These paths represent the most promising graph evidence, but their relations and entities still need to be interpreted together with the image before we can determine whether they are sufficient to answer $Q$. Therefore, this module verifies the retained paths against the visual content and decides whether the current multimodal evidence can support answer generation or whether the search should continue to the next hop.

We first summarize the retained paths under the visual context. Given the image $I$, the question $Q$, and the retained path set $\mathcal{P}^{(i)}$, we use a vision--language model $\texttt{VLM}(\cdot)$ to generate an image-grounded evidence statement:
\begin{equation}
\label{eq:evidence-summary}
z_i =
\texttt{VLM}
\left(
I, Q, \mathcal{P}^{(i)}
\right),
\end{equation}
where $z_i$ summarizes how the selected graph paths relate to the visual content. This step is necessary because the retained paths describe symbolic graph relations and entities, while the final answer should remain consistent with the depicted object and the visual evidence in $I$.

We then query $\texttt{VLM}(\cdot)$ to assess whether the image-grounded evidence statement $z_i$ is sufficient to answer $Q$. The model returns an answerability verdict and, if the verdict is positive, a candidate answer $A_i$. We formulate this step as
\begin{equation}
\label{eq:answerability}
(\eta_i, A_i)=\texttt{VLM}\left(I,Q,z_i\right),
\end{equation}
where $\eta_i\in\{0,1\}$ indicates whether the current multimodal evidence is sufficient to answer $Q$, and $A_i$ denotes the candidate answer generated from $z_i$ when $\eta_i=1$. \methodname{} stops normal path expansion only when the evidence is judged answerable and the predicted reasoning depth $\widehat{D}$ has been reached. Accordingly, we define the stopping hop as
\begin{equation}
\label{eq:stop-hop}
i_{\mathrm{stop}}=\min\left\{
i \mid \eta_i=1 \ \wedge\ i\geq \widehat{D}
\right\}.
\end{equation}
If such a hop exists, the final answer is generated from the corresponding image-grounded evidence. If the evidence becomes answerable before reaching $\widehat{D}$, we record the candidate answer as a fallback but continue the search. This prevents a shallow path from prematurely terminating the reasoning process before the expected number of hops is completed. 

When neither a valid stopping hop nor a fallback hop exists, \methodname{} performs terminal answer selection. This step ranks the endpoint entities appearing in the retained paths and selects the most plausible answer candidate. When reference images are available, the top candidates are further verified against $I$ using $\texttt{VLM}(\cdot)$. If no reliable graph candidate can be obtained, the answer is inferred directly from the image. In this way, \methodname{} keeps visual evidence involved until the final answer is produced, while avoiding unnecessary hops once sufficient multimodal evidence has been obtained.

\input{algorithm/reasoning}

%% file: fig/framework.tex
\begin{figure*}[t!]
    \centering
    \includegraphics[scale=0.65]{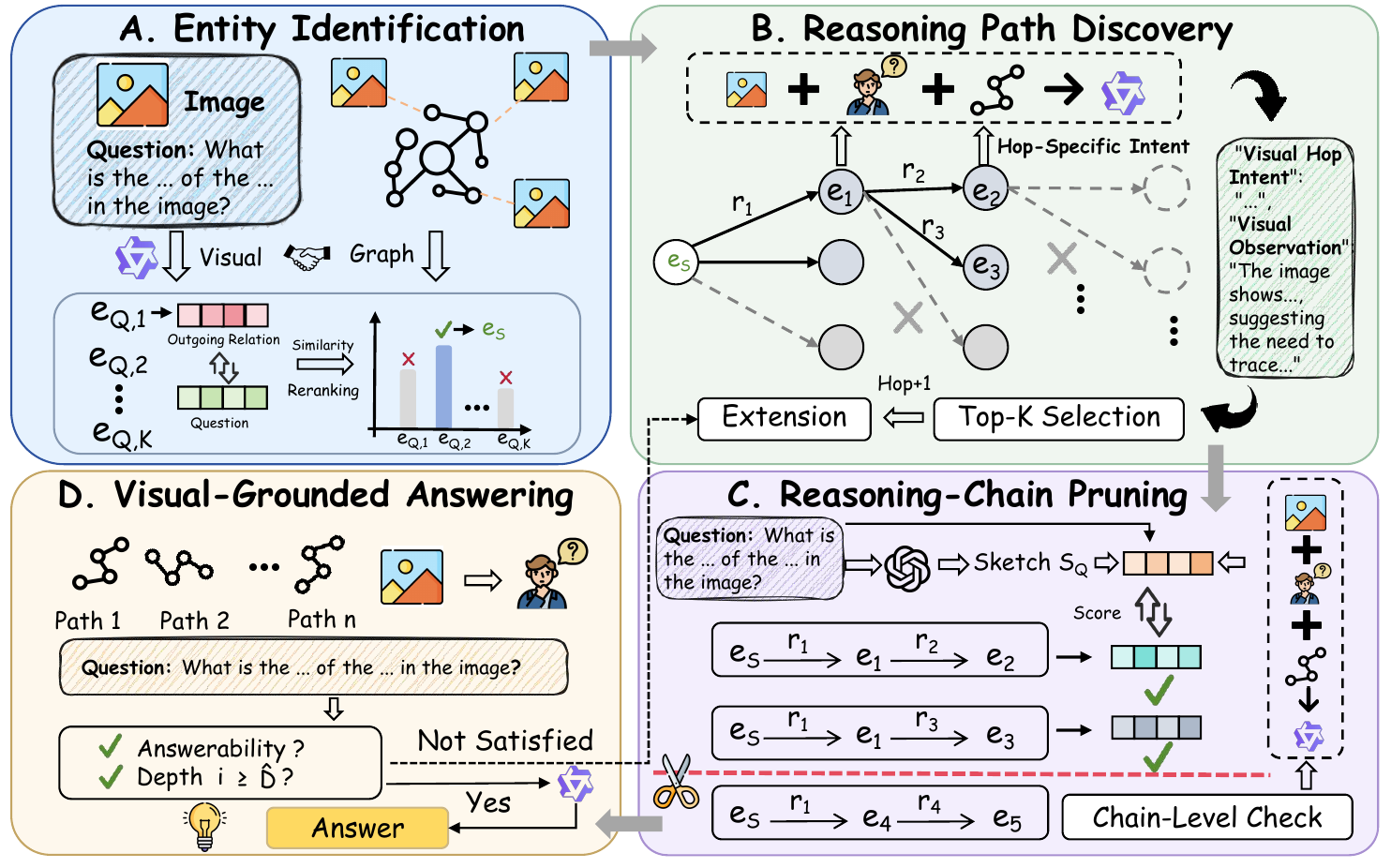}
\caption{Framework of the proposed \methodname{}. \methodname{} answers queries through four coordinated stages: (a) visual- and graph-aware entity identification that selects a structurally valid starting entity from image-related candidates, (b) intent-guided reasoning path discovery that uses hop-specific visual intent to guide relation expansion and Top-K selection, (c) reasoning-chain pruning that evaluates candidate paths with respect to the question sketch and chain-level coherence, and (d) visual-grounded answering that checks answerability with the image and selected paths before producing the final answer.}
    \label{fig:framework}
\end{figure*}

%% file: algorithm/reasoning.tex
\begin{algorithm}[t]
\caption{\methodname{} Reasoning Process}
\label{alg:vispath}
\small
\DontPrintSemicolon
\SetKwComment{Comment}{$\triangleright$\ }{}
\SetKwInOut{Input}{Input}
\SetKwInOut{Output}{Output}
\Input{image $I$, question $Q$, MMKG $\mathcal{G}$, retention sizes $M_1$, $M_2$, $M_3$, depth budget $D_{\max}$}
\Output{answer $A$}
recognize object $o$ from $I$ and rank entity candidates $\{e_{Q,1},\dots,e_{Q,K}\}$\Comment*[r]{Eq.~\eqref{eq:visual-recog}: visual recognition}
select the starting entity $e_S$ by graph-aware validation\Comment*[r]{Eqs.~\eqref{eq:edge-align}--\eqref{eq:graph-aware-correction}}
prompt the LLM for the reasoning sketch $\mathcal{S}_Q$ and predicted depth $\widehat{D}$; \quad $\widehat{D} \leftarrow \min(\widehat{D},\, D_{\max})$\;
$\mathcal{P}^{(0)} \leftarrow \{(e_S)\}$;\quad $z^\star \leftarrow \bot$\;
\For{$i = 1$ \KwTo $D_{\max}$}{
  $q_Q^{(i)} \leftarrow \texttt{VLM}(I, Q, \mathcal{P}^{(i-1)})$\Comment*[r]{Eq.~\eqref{eq:hop-intent}: hop intent}
  keep the top-$M_1$ intent-matched edges per frontier entity\Comment*[r]{Eq.~\eqref{eq:edge-score}: edge scoring}
  extend $\mathcal{P}^{(i-1)}$ with retained edges into $\mathcal{P}_{\mathrm{cand}}^{(i)}$\Comment*[r]{Eq.~\eqref{eq:path-extension}: path extension}
  retain the top-$M_2$ paths as $\mathcal{P}_{\mathrm{chain}}^{(i)}$ by chain scores $\sigma_{p,i}$\Comment*[r]{Eq.~\eqref{eq:path-rerank}: chain reranking}
  examine $\mathcal{P}_{\mathrm{chain}}^{(i)}$ with the LLM and retain at most $M_3$ paths as $\mathcal{P}^{(i)}$\;
  $z_i \leftarrow \texttt{VLM}(I, Q, \mathcal{P}^{(i)})$\Comment*[r]{Eq.~\eqref{eq:evidence-summary}: evidence statement}
  \If{$z_i$ supports answering $Q$}{
    $z^\star \leftarrow z_i$\Comment*[r]{Eq.~\eqref{eq:answerability}: answerability}
    \If{$i \geq \widehat{D}$}{\textbf{break}\Comment*[r]{Eq.~\eqref{eq:stop-hop}: stop rule}}
  }
}
\eIf{$z^\star \neq \bot$}{
  generate the final answer $A$ from $(I, Q, z^\star)$\;
}{
  select $A$ from visited path entities with visual verification\;
}
\Return{$A$}
\end{algorithm}

%% file: sec/5_experiments.tex
\section{Experiments}
\label{sec:experiments}

\subsection{\methodname{}-Bench construction}
\begin{figure}[t!]
    \centering
    \includegraphics[scale=0.55]{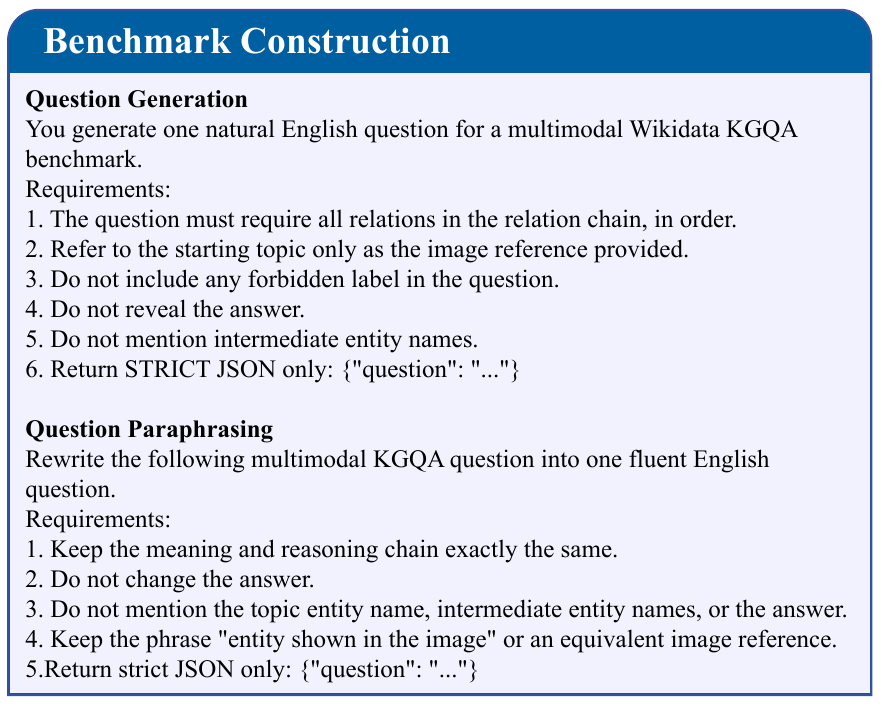}
    \caption{Prompt templates used in the construction of \methodname{}-Bench.}
    \label{fig:prompt-benchmark}
\end{figure}

Existing MMKGQA benchmarks are dominated by two-hop questions and contain few examples that require deeper reasoning chains. To evaluate multimodal multi-hop reasoning more systematically, we construct \methodname{}-Bench from the Wikidata5M MMKG in three stages. (i)~\emph{Sampling}: we sample topic entities that are associated with an image through the Wikidata \texttt{P18} property and have gold reasoning paths of length two to four. (ii)~\emph{Question generation}: for each sampled path, we generate a natural-language question whose answer is the path endpoint. The topic entity is referred to only through a visual referring expression, so that the starting entity must be grounded from the image rather than read directly from the question text. (iii)~\emph{Single-answer filtering}: we retain only questions whose gold path leads to a single unambiguous answer entity, and discard paths whose relations allow multiple valid endpoints. The resulting benchmark contains 2{,}000 questions, including 500 two-hop, 1{,}200 three-hop, and 300 four-hop questions, covering 52 distinct Wikidata relations. We provide the prompt templates in Figure~\ref{fig:prompt-benchmark} and distribution of the question length in Figure~\ref{fig:question_len}.

\input{tables/dataset_stats}

\subsection{Experimental Setup}
\label{sec:exp-setup}
\noindent\textbf{Benchmarks.} We evaluate \methodname{} on four multi-hop visual question answering benchmarks, as summarized in Table~\ref{tab:dataset_stats}. These include our \methodname{}-Bench, described below, and three benchmarks adapted from prior work. ReasonVQA~\cite{tran2025reasonvqa} and MMhops~\cite{zhang2026mmhops} are entity-centric multi-hop VQA datasets. To make them compatible with our path-based evaluation protocol, we re-ground each question onto the knowledge graph by resolving its topic entity and gold answer to KG nodes. E-VQA~\cite{mensink2023encyclopedic} is a two-hop encyclopedic VQA dataset built over fine-grained natural and landmark images. Together, the four benchmarks cover two to four reasoning hops and involve diverse image sources, including Visual Genome, Wikipedia, iNaturalist, Google Landmarks, and the Wikidata imagery used in our own benchmark.

\noindent\textbf{Baselines.} We compare \methodname{} with four families of baselines: (i)~classic KGQA methods, including EmbedKGQA~\cite{saxena2020embedkgqa}; (ii)~vision--language models evaluated zero-shot on $(I,Q)$, including open-source models Qwen2.5-VL-32B/72B~\cite{bai2025qwen25vl} and GLM-4.6V~\cite{glmv2025}, as well as closed-source models GPT-4o~\cite{hurst2024gpt4o} and GPT-5.4; (iii)~multi-hop KG reasoning methods, including MindMap~\cite{wen2024mindmap}, ToG~\cite{sun2024tog}, and PoG~\cite{tan2025paths}; and (iv)~retrieval-augmented and agentic methods, including OmniSearch~\cite{li2025OmniSearch} and MMSearch-R1~\cite{wu2025mmsearch}. The text-only KGQA methods in family~(iii) cannot access the image and therefore require a starting entity. For a controlled comparison, we provide these text-only KGQA baselines with the gold starting entity, since they cannot access the image for entity grounding. In contrast, \methodname{} must infer the starting entity from the multimodal input before performing graph reasoning. Thus, the observed improvements are achieved under a stricter setting for \methodname{}, further demonstrating the effectiveness of incorporating multimodal evidence into hop-by-hop path reasoning.

\input{fig/question_len_fig}

\noindent\textbf{Metrics.} Since the answers are free-form text and may involve aliases, hypernyms, or multilingual variants, exact match can underestimate model performance and may unfairly penalize verbose outputs from VLM baselines. We therefore report open-ended semantic accuracy, denoted as SBERT@0.3, computed with Sentence-BERT embeddings~\cite{reimers2019sentencebert}. A prediction is considered correct if the cosine similarity between its sentence embedding and that of the gold answer is at least $0.3$. We report both per-hop and overall accuracy, with all results micro-averaged over questions. The similarity threshold follows the open-ended evaluation protocol used in prior multimodal QA studies~\cite{mensink2023encyclopedic,tran2025reasonvqa}.

\noindent\textbf{Implementation details.} \methodname{} uses GPT-4o as the text language model, Qwen2.5-VL-72B as the vision--language model, and all-MiniLM-L6-v2~\cite{reimers2019sentencebert} as the sentence encoder. We adopt Wikidata5M~\cite{wang2021kepler} as the underlying knowledge graph and build an approximate nearest-neighbor index over its entity names for efficient entity retrieval. The main hyperparameters are the per-hop edge retention size $M_1=30$ for path expansion, the path reranking size $M_2=30$ for chain-level scoring, the chain retention size $M_3=3$ for beam-style selection, the override threshold $\delta{=}0.05$ in Eq.~\eqref{eq:graph-aware-correction}, which grows by $0.01$ per rank for lower-ranked candidates to keep overrides conservative, and the maximum search depth $D_{\max}=3$.

\subsection{Main Results}
\label{sec:exp-main}
\input{tables/main_results}

We evaluate \methodname{} against a comprehensive set of baselines on four benchmarks under the SBERT@0.3 metric, as shown in Table~\ref{tab:main_results}. The results first show that \methodname{}-Bench is more challenging than existing benchmarks for current baselines. For example, strong zero-shot VLMs suffer clear performance drops on \methodname{}-Bench compared with other datasets: Qwen2.5-VL-72B drops by $26.4\%$ relative to its performance on MMhops, and GLM-4.6V drops by $37.0\%$. This indicates that recognizing the depicted entity or relying on dense encyclopedic knowledge is insufficient for our benchmark, where the answer must be derived through image-grounded multi-hop paths.

The results also show that reasoning depth substantially affects all methods. As the number of hops increases, text-only path-reasoning methods degrade sharply because they decode relation sequences mainly from the question and cannot revise them with multimodal evidence. For instance, PoG drops by $54.2\%$ from two-hop to four-hop questions on \methodname{}-Bench, while ToG drops by $59.7\%$. Even strong VLM baselines show notable degradation as hop depth increases, suggesting that implicit reasoning from the input alone is not enough for deep MM-KGQA.

In contrast, \methodname{} delivers more uniform performance across benchmarks and reasoning depths. On \methodname{}-Bench, \methodname{} achieves a $10.6\%$ relative improvement in average accuracy over the strongest baseline. Its advantage holds consistently across different hop levels, with relative improvements of $13.1\%$, $9.9\%$, and $8.2\%$ at two, three, and four hops, respectively. These results demonstrate that keeping multimodal evidence involved in hop-by-hop path discovery and reasoning-chain selection enables \methodname{} to remain robust as questions become deeper and more compositional.

\subsection{Backbone Study}
\label{sec:exp-backbone}
Tables~\ref{tab:text_backbone} and~\ref{tab:vision_backbone} evaluate the robustness of \methodname{} under different text and vision backbones. For the text LLM, GPT-4o and GPT-4o-mini achieve comparable performance on \methodname{}-Bench, suggesting that our path reasoning framework does not rely solely on the strongest proprietary text model. However, the gap becomes clearer on ReasonVQA, where GPT-4o provides stronger cross-benchmark generalization. DeepSeek-V3 and Qwen3-32B remain competitive on two-hop questions, but their performance drops more substantially at deeper hops, indicating that long-range path planning and reasoning-chain selection require stronger instruction following and semantic judgment. For the vision backbone, Qwen2.5-VL-72B achieves the best overall performance across both benchmarks. Qwen3-VL-32B remains competitive on two-hop questions but degrades more on deeper reasoning, while Qwen3-VL-8B shows a larger drop, especially on \methodname{}-Bench. These results show that both components are important: the vision model affects multimodal grounding and hop-specific intent estimation, while the text LLM is crucial for maintaining coherent multi-hop reasoning and selecting globally valid evidence paths.

\input{tables/backbone}

\subsection{Topic Grounding Analysis}
\label{sec:exp-grounding}
% Claims: grounding is the bottleneck; self-correcting validation improves it;
% \methodname{} tolerates an imperfect starting entity (Table~\ref{tab:grounding_tolerance}).
\input{tables/grounding}

Table~\ref{tab:grounding} evaluates starting-entity grounding on ReasonVQA. Here, \emph{Top-1 Selection} denotes a direct grounding strategy that simply takes the highest-ranked entity predicted by the vision-language model, without applying graph-aware self-correcting validation. This setting measures how reliable visual-name matching alone is for recovering the correct starting entity. As shown in the table, Top-1 Selection with Qwen2.5-VL-72B achieves only $24.1$ strict accuracy and $25.1$ soft accuracy, indicating that grounding the starting entity from the multimodal input is a non-trivial bottleneck. When applying the self-correcting validation in \methodname{} under the same Qwen2.5-VL-72B backbone, strict accuracy increases to $42.7$ and soft accuracy increases to $48.3$, corresponding to an $18.6$-point strict gain and a $23.2$-point soft gain. This confirms that visual similarity alone is insufficient, while checking whether candidate entities expose question-relevant outgoing relations can substantially improve grounding reliability. The results also show that grounding performance varies across vision backbones: replacing Qwen2.5-VL-72B with GPT-4o further improves strict accuracy to $56.2$, and GLM-4.6V achieves the best performance with $57.2$ strict accuracy and $64.8$ soft accuracy. These results demonstrate that both multimodal recognition quality and graph-aware validation are important for accurate starting-entity grounding, and they also explain why robust downstream path reasoning is necessary when the initial grounding remains imperfect.

\subsection{Ablation Study}
\label{sec:exp-ablation}
% Claim: each of the four components contributes; removing any one drops accuracy.
\input{tables/ablation}

Table~\ref{tab:ablation} reports the component ablation results on \methodname{}-Bench, where one component is removed from the full method at a time. Removing \emph{Graph-aware validation} disables the relation-compatibility correction in Eq.~\eqref{eq:graph-aware-correction}, so the starting entity is directly selected from the top-ranked multimodal grounding candidate. Removing \emph{Visual intent} discards the hop-specific intent $q_Q^{(i)}$ generated by Eq.~\eqref{eq:hop-intent}; consequently, outgoing edges in Eq.~\eqref{eq:edge-score} are matched only against the question text during path discovery in Section~\ref{sec:method-discovery}. Removing \emph{Reasoning-chain pruning} eliminates the chain-level scoring in Eq.~\eqref{eq:path-rerank} and the subsequent LLM-based chain examination in Section~\ref{sec:method-selection}, so candidate paths are retained solely according to hop-level edge scores. Removing \emph{Evidence checking} disables the answerability verification in Eq.~\eqref{eq:answerability} and the stopping rule in Eq.~\eqref{eq:stop-hop}, forcing the search to run until the full depth budget is reached before answer generation.

The results support two main observations. First, all components contribute to the overall performance, with graph-aware validation being the most critical one. Removing this component decreases the average score from $64.5$ to $56.2$ and consistently degrades performance at all hop depths. This finding is consistent with the grounding analysis in Section~\ref{sec:exp-grounding}: once an incorrect starting entity is selected, later reasoning stages have limited ability to recover the correct answer path. Removing visual intent and reasoning-chain pruning also leads to clear performance drops, reducing the average score by $5.5$ and $6.0$ points, respectively. This shows that hop-level multimodal guidance and chain-level consistency filtering address complementary failure modes: the former helps select relevant relations at each step, while the latter prevents locally plausible edges from forming globally inconsistent paths. Second, evidence checking has a different role from the other components. Removing it leaves the 2-hop score unchanged, decreases the 3-hop score from $62.4$ to $61.6$, and slightly increases the 4-hop score from $57.0$ to $58.0$. This is because the answerability gate may occasionally be over-confident on deeper questions and stop the search with an answer supported by a partial evidence chain. Without this gate, the model always exhausts the depth budget and may therefore reach the deeper evidence required by some 4-hop questions. However, for shallower questions, evidence checking helps avoid unnecessary expansion beyond an already sufficient answer path, which explains the drop when it is removed. Overall, the full model achieves the best average performance while also avoiding unnecessary deeper exploration, showing that evidence checking provides a useful balance between answer quality and search efficiency.

\subsection{Cost and Efficiency}
\label{sec:exp-efficiency}

\input{tables/cost}

Table~\ref{tab:cost} reports the per-question cost of \methodname{} and representative baselines on \methodname{}-Bench. Although \methodname{} introduces additional VLM calls to keep multimodal evidence involved during hop-by-hop reasoning, its overall cost remains controllable. With GPT-4o-mini, \methodname{} uses $5.8$ LLM calls and $5.2$ VLM calls per question, while consuming only $3.6$k text tokens, which is lower than ToG and OmniSearch and comparable to PoG. Its wall-clock time is also close to existing multi-hop and retrieval-augmented baselines, showing that the additional multimodal reasoning steps do not lead to prohibitive overhead. Using GPT-4o increases the runtime because of the stronger and slower backbone, but the number of calls and token usage remain within a moderate range. These results indicate that \methodname{} achieves stronger multimodal multi-hop reasoning while maintaining a practical and manageable inference cost.

\subsection{Hyperparameter Study}
\label{sec:exp-hyperparam}
\input{fig/hyperparam_fig}

\begin{figure*}[t!]
  \centering
   \includegraphics[width=\linewidth]{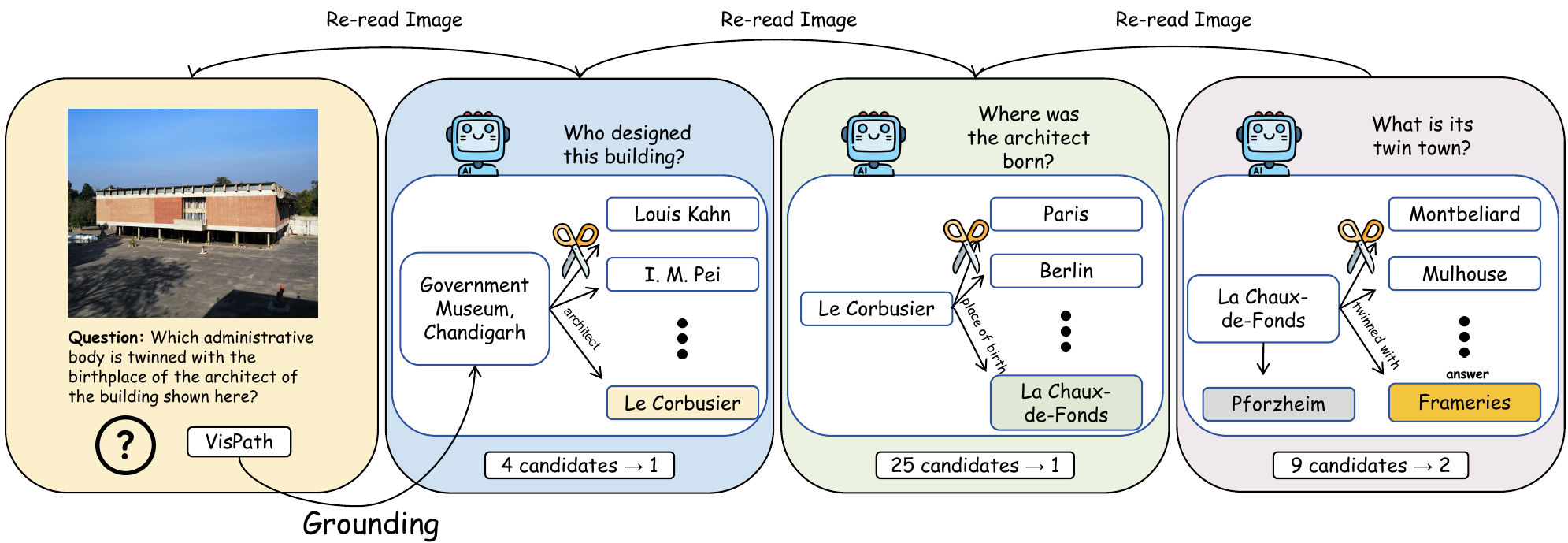}
  \caption{Qualitative example: the candidate paths \methodname{} enumerates and the path it selects to answer the query.}
  \label{fig:case_study}
\end{figure*}

\begin{figure}[t!]
    \centering
    \includegraphics[scale=0.55]{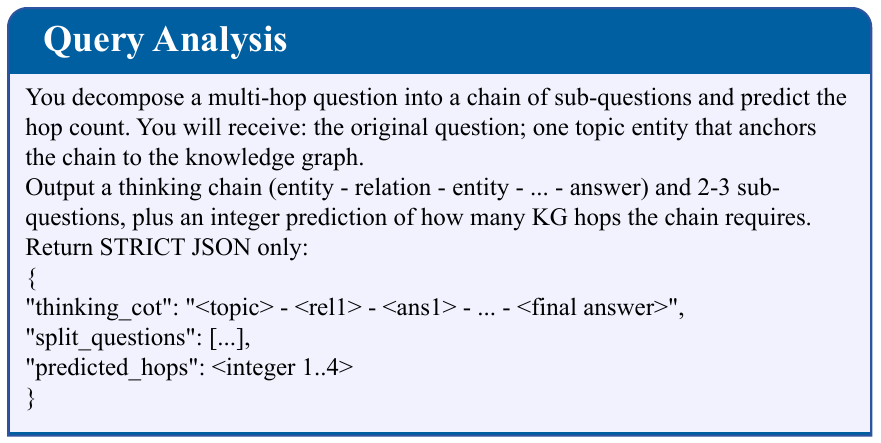}
    \caption{Prompt template of question analysis.}
    \label{fig:prompt_analysis}
\end{figure}

\begin{figure}[t!]
    \centering
    \includegraphics[scale=0.55]{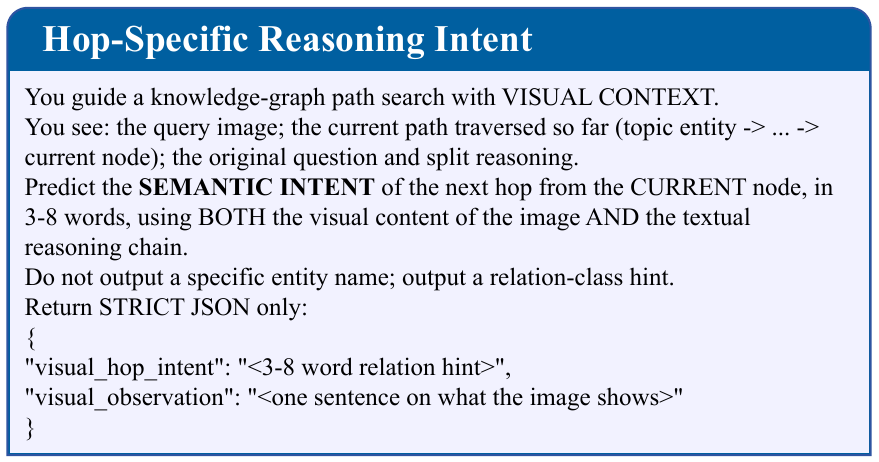}
    \caption{Prompt template of intent generation.}
    \label{fig:prompt_intent}
\end{figure}

\begin{figure}[t!]
    \centering
    \includegraphics[scale=0.55]{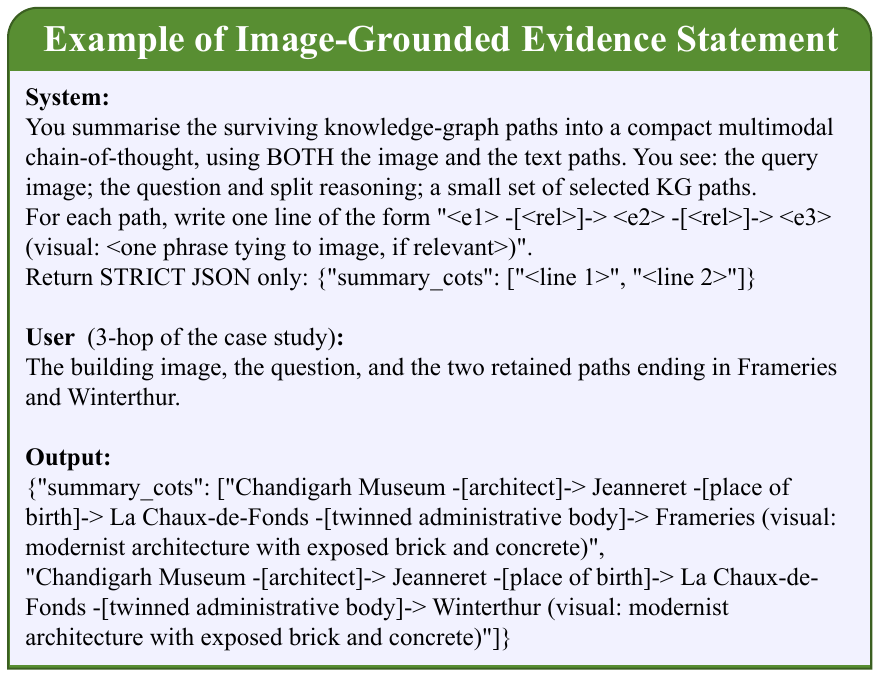}
    \caption{Example of generating statement.}
    \label{fig:prompt_statement}
\end{figure}

\begin{figure}[t!]
    \centering
    \includegraphics[scale=0.55]{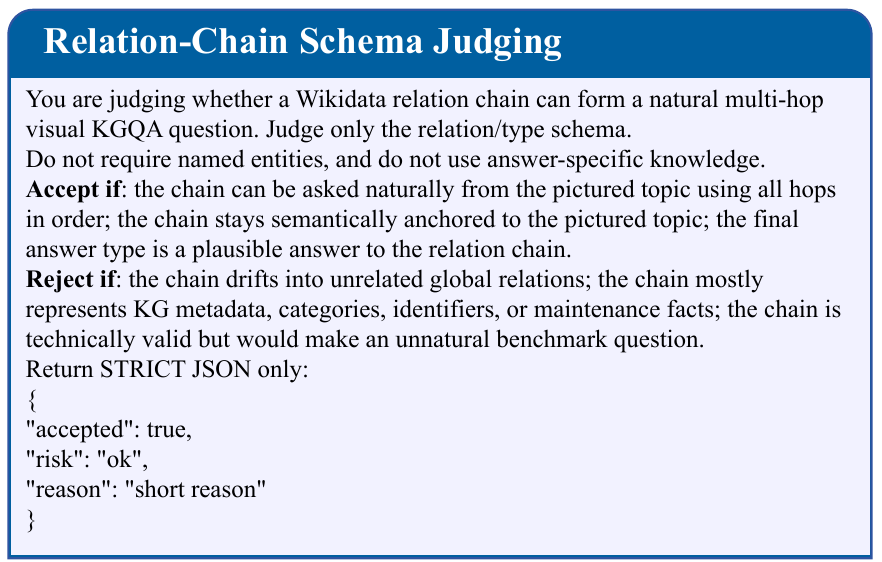}
    \caption{Prompt template for relation-chain schema judging}
    \label{fig:prompt_check}
\end{figure}

Figure~\ref{fig:hyperparam} studies the sensitivity of \methodname{} to three key hyperparameters: the per-hop edge retention size $M_1$ from Section~\ref{sec:method-discovery}, the chain retention size $M_3$ from Section~\ref{sec:method-selection}, and the maximum search depth $D_{\max}$, which truncates the predicted depth $\widehat{D}$ from the query analysis when the prediction is overly aggressive. Overall, \methodname{} is not highly sensitive to these settings. Increasing $M_1$ from $15$ to $30$ improves the average accuracy by covering more useful outgoing edges at each hop, while further increasing it to $60$ brings no additional gain, suggesting that a moderate retention size already covers the relevant candidates and larger values mainly admit irrelevant ones. For the chain retention size, performance increases substantially from $M_3{=}1$ to $M_3{=}3$, because keeping a single chain makes the search fragile to one wrong selection, while a small set of globally coherent chains provides sufficient redundancy. Further increasing $M_3$ to $5$ leaves the result nearly unchanged. The depth budget has the smallest impact, as the predicted depth $\widehat{D}$ rarely exceeds the actual hop count of the benchmark questions, so enlarging $D_{\max}$ changes the executed search for only a small fraction of queries. These observations show that \methodname{} is robust to hyperparameter choices, and the default setting of $M_1{=}30$, $M_3{=}3$, and $D_{\max}{=}3$ provides a good balance between reasoning coverage and cost.

\subsection{Case Study}
\label{sec:exp-case}
Figure~\ref{fig:case_study} illustrates a complete reasoning process of \methodname{} on a representative query. Given the input image and the question ``Which administrative body is twinned with the birthplace of the architect of the building shown here?'', \methodname{} first grounds the visual content to the starting entity \textit{Government Museum, Chandigarh}. It then performs three-hop reasoning, where each hop is guided by a multimodal intent.

At the first hop, \methodname{} identifies that the question requires finding the architect of the depicted building. Based on the visual appearance of the building and the current reasoning state, it expands candidate relations from the starting entity and selects the path leading to \textit{Le Corbusier}. At the second hop, the reasoning intent shifts to the architect's birthplace. Among the candidate paths expanded from \textit{Le Corbusier}, \methodname{} selects the path reaching \textit{La Chaux-de-Fonds}. At the third hop, the intent further changes to the administrative body twinned with that birthplace. The model reranks the candidate continuations and retains the path that leads to \textit{Frameries}, which is returned as the final answer.

This example shows how \methodname{} keeps multimodal evidence involved throughout the reasoning process rather than using it only for initial grounding. The same input image supports different reasoning intents at different hops, including identifying the building's architect, locating the architect's birthplace, and finding the corresponding twinned administrative body. In contrast, a text-only method must infer the entire relation sequence from the question alone and cannot adjust its path search using multimodal evidence at intermediate steps. By repeatedly re-anchoring the search to the multimodal input and selecting paths at the reasoning-chain level, \methodname{} maintains a coherent evidence path from the starting entity to the final answer.

\subsection{Prompt Templates}

\label{sec:exp-prompts}

To improve the reproducibility of \methodname{}, we present the key prompt templates used in our reasoning pipeline. Specifically, Figure~\ref{fig:prompt_analysis} shows the query analysis prompt, which decomposes the input question into a reasoning sketch $\mathcal{S}_Q$ and predicts the required reasoning depth $\widehat{D}$. Figure~\ref{fig:prompt_intent} presents the hop-specific intent generation prompt, which produces the visual reasoning intent $q_Q^{(i)}$ at each hop based on the image, the question, and the current partial path. Figure~\ref{fig:prompt_statement} illustrates the prompt for generating the image-grounded evidence statement $z_i$, where selected KG paths are summarized together with the visual observation. All these prompts require strict JSON outputs to ensure stable parsing and controllable downstream execution. In addition, we provide the prompt templates used for relation-chain schema judging in Figure~\ref{fig:prompt_check}.

%% file: tables/dataset_stats.tex
\begin{table*}[t]
\centering
\footnotesize
\setlength{\tabcolsep}{4pt}
\renewcommand{\arraystretch}{1.15}
\caption{Statistics of the four multi-hop VQA benchmarks. Avg.\ $|Q|$ and Avg.\ $|A|$ denote the average question and answer lengths in words, respectively. \#Relations reports the number of distinct relations appearing in the gold reasoning paths.}
\label{tab:dataset_stats}
\begin{tabular}{l|c|ccc|cccc}
\toprule
\multirow{2}{*}{\textbf{Dataset}} & \multirow{2}{*}{\textbf{\#Questions}}
& \multicolumn{3}{c|}{\textbf{Hop Distribution}}
& \multirow{2}{*}{\textbf{Avg.\ $|Q|$}} & \multirow{2}{*}{\textbf{\#Relations}} & \multirow{2}{*}{\textbf{Avg.\ $|A|$}} & \multirow{2}{*}{\textbf{Image Source}} \\
& & 2-Hop & 3-Hop & 4-Hop & & & & \\
\midrule
ReasonVQA        & 1{,}951 & 367   & 1{,}584 & --  & 13.8 & 49   & 1.8 & Visual Genome \\
MMhops                      & 1{,}907 & 874   & 829     & 204 & 15.6 & --  & 1.7 & Wikipedia \\
E-VQA                       & 1{,}400 & 1{,}400 & --    & --  & 16.6 & --  & 1.7 & iNat/GLDv2 \\
\midrule
\methodname{}-Bench & 2{,}000 & 500   & 1{,}200 & 300 & 18.4 & 52   & 2.3 & Wikidata (P18) \\
\bottomrule
\end{tabular}
\end{table*}

%% file: fig/question_len_fig.tex
% §5.1 benchmark 处或附录: \input{figs/question_len_fig}
\begin{figure}[t!]
  \centering
  \includegraphics[width=\linewidth]{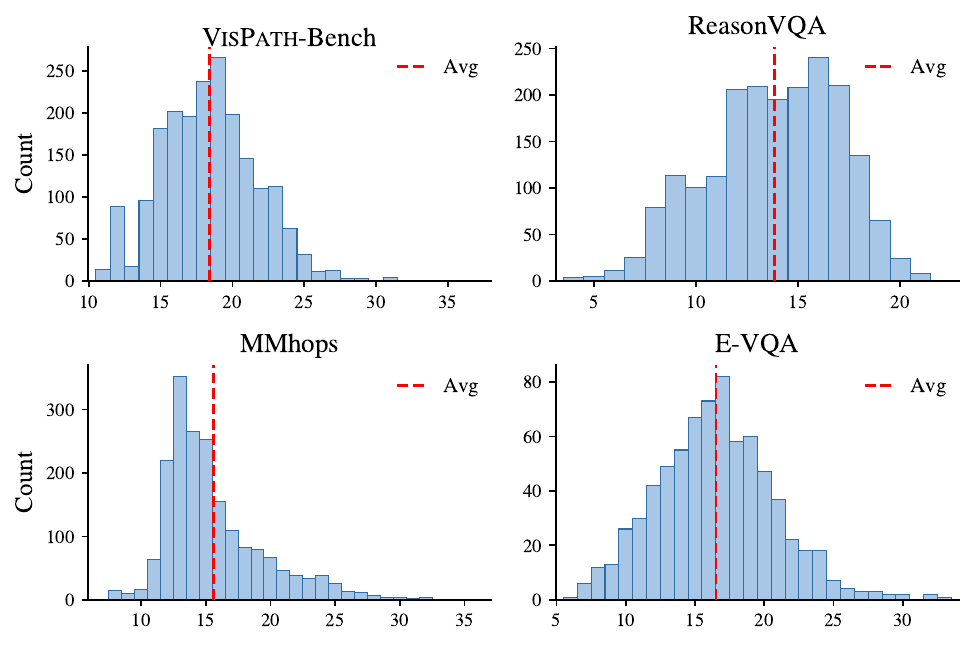}
  \caption{Distribution of question lengths on the four benchmarks.}
  \label{fig:question_len}
\end{figure}

%% file: tables/main_results.tex
\begin{table*}[t]
\centering
\footnotesize
\setlength{\tabcolsep}{3.8pt}
\renewcommand{\arraystretch}{1.15}
\caption{Main results on four multi-hop VQA benchmarks. Metric: SBERT@0.3.
\textbf{Bold}: best in column; \underline{underline}: second-best.}
\label{tab:main_results}
\begin{tabular}{l|cccc|ccc|cccc|c}
\toprule
\multirow{2}{*}{\textbf{Method}}
& \multicolumn{4}{c|}{\textbf{\methodname{}-Bench}}
& \multicolumn{3}{c|}{\textbf{ReasonVQA}}
& \multicolumn{4}{c|}{\textbf{MMhops}}
& \textbf{E-VQA} \\
& 2-hop & 3-hop & 4-hop & Avg
& 2-hop & 3-hop & Avg
& 2-hop & 3-hop & 4-hop & Avg
& 2-hop \\
\midrule
\multicolumn{13}{c}{\cellcolor{gray!10}\emph{ Classic KGQA}} \\
EmbedKGQA \scriptsize{(ACL'20)}
& 16.2 & 11.1 & \phantom{0}4.7 & 11.4
& 28.6 & 15.9 & 18.3
& \phantom{0}9.8 & \phantom{0}6.9 & \phantom{0}7.4 & \phantom{0}8.3
& \phantom{0}5.9 \\
\midrule
\multicolumn{13}{c}{\cellcolor{gray!10}\emph{Open-source Vision–Language Models}} \\
Qwen2.5-VL-32B \scriptsize{(2025)}
& 58.6 & 49.5 & 46.0 & 51.3
& 49.2 & 51.4 & 51.0  
& 70.6 & 67.8 & 70.2  & 69.3
& 61.7 \\
Qwen2.5-VL-72B \scriptsize{(2025)}
& 62.0 & 52.6 & 49.7 & 54.5
& 46.6 & 57.1 & 55.1
& 75.4 & 72.4 & 75.0 & 74.0
& 65.0 \\
GLM-4.6V \scriptsize{(2025)}
& 51.8 & 50.0 & 48.3 & 50.2
& 52.0 & 60.5 & 58.9
& 80.3 & \underline{79.6} & \underline{77.9} & 79.7
& \underline{66.4} \\
\midrule
\multicolumn{13}{c}{\cellcolor{gray!10}\emph{Closed-source Vision–Language Models}} \\
GPT-4o \scriptsize{(2024)}
& 57.6 & 50.1 & 47.0 & 51.5
& 63.1 & 52.8 & 54.9
& 79.1 & 74.3 & 77.5 & 76.8
& 62.8 \\
GPT-5.4 \scriptsize{(2026)}
& 65.4 & \underline{56.8} & \underline{52.7} & \underline{58.3}
& \underline{69.6} & \underline{60.7} & \underline{62.4}
& \underline{81.7} & 78.9 & 76.0 & \underline{79.9}
& 66.0 \\  
\midrule
\multicolumn{13}{c}{\cellcolor{gray!10}\emph{Multi-Hop Reasoning}} \\
MindMap \scriptsize{(ACL'24)}
& \underline{68.4} & 50.6 & 49.7 & 54.9
& 51.2 & 47.6 & 48.3
& 52.4 & 52.4 & 46.1 & 51.6
& 41.0 \\
ToG (GPT-4o) \scriptsize{(ICLR'24)}
& 34.0 & 16.7 & 13.7 & 20.6
& 32.4 & 22.1 & 24.0
& 27.6 & 22.7 & 17.2 & 24.3
& 11.4 \\
PoG (GPT-4o) \scriptsize{(WWW'25)}
& 55.2 & 36.7 & 25.3 & 39.6
& 31.6 & 32.1 & 32.0
& 34.1 & 26.0 & 16.7 & 29.4  
& 17.1 \\
\midrule
\multicolumn{13}{c}{\cellcolor{gray!10}\emph{ Retrieval-Augmented / Agent}} \\
OmniSearch \scriptsize{(ICLR'25)}
& 38.0 & 36.6 & 37.0 & 37.0
& 37.9 & 43.1 & 42.1
& 50.7 & 38.6 & 28.4 & 43.1
& 37.4 \\  
MMSearch-R1-7B \scriptsize{(ACL'26)}
& 44.2 & 38.1 & 35.0 & 39.2
& 45.5 & 51.1 & 50.0
& 60.9 & 48.0 & 43.1 & 53.3
& 32.3 \\ 
\midrule
\rowcolor{blue!8}
\textbf{\methodname{} (GPT-4o)}
& \textbf{74.0} & \textbf{62.4} & \textbf{57.0} & \textbf{64.5}
& \textbf{81.5} & \textbf{64.8} & \textbf{68.0}  
& \textbf{86.5} & \textbf{83.1} &  \textbf{79.8}  & \textbf{84.3}
& \textbf{70.4} \\
\bottomrule
\end{tabular}
\end{table*}

%% file: tables/backbone.tex
\begin{table}[t!]
\centering
\footnotesize
\setlength{\tabcolsep}{4pt}
\renewcommand{\arraystretch}{1.15}
\caption{Performance of \methodname{} with different text LLM backbones. The vision model is fixed to Qwen2.5-VL-72B.}
\label{tab:text_backbone}
\begin{tabular}{l|cccc|ccc}
\toprule
\multirow{2}{*}{\textbf{Text LLM}}
& \multicolumn{4}{c|}{\textbf{\methodname{}-Bench}}
& \multicolumn{3}{c}{\textbf{ReasonVQA}} \\
& 2-hop & 3-hop & 4-hop & Avg
& 2-hop & 3-hop & Avg \\
\midrule
GPT-4o      & 74.0 & 62.4 & 57.0 & 64.5 & 81.5 & 64.8 & 68.0 \\
GPT-4o-mini & 74.2 & 61.3 & 56.4 & 63.8 & 75.9 & 53.3 & 57.6 \\
DeepSeek-V3 & 73.2 & 40.7 & 33.0 & 47.7 & 66.7 & 55.7 & 57.8 \\
Qwen3-32B   & 74.4 & 59.3 & 47.3 & 61.3 & 55.2 & 47.9 & 49.3 \\
\bottomrule
\end{tabular}
\end{table}

\begin{table}[t!]
\centering
\footnotesize
\setlength{\tabcolsep}{4pt}
\renewcommand{\arraystretch}{1.15}
\caption{Performance of \methodname{} with different vision model backbones. The text LLM is fixed to GPT-4o.}
\label{tab:vision_backbone}
\begin{tabular}{l|cccc|ccc}
\toprule
\multirow{2}{*}{\textbf{Vision Model}}
& \multicolumn{4}{c|}{\textbf{\methodname{}-Bench}}
& \multicolumn{3}{c}{\textbf{ReasonVQA}} \\
& 2-hop & 3-hop & 4-hop & Avg
& 2-hop & 3-hop & Avg \\
\midrule
Qwen2.5-VL-72B & 74.0 & 62.4 & 57.0 & 64.5 & 81.5 & 64.8 & 68.0 \\
Qwen3-VL-32B   & 75.6 & 58.6 & 51.7 & 61.8 & 76.4 & 61.9 & 64.7 \\
Qwen3-VL-8B    & 63.0 & 53.2 & 45.0 & 54.4 & 52.9 & 61.4 & 59.8 \\
\bottomrule
\end{tabular}
\end{table}

%% file: tables/grounding.tex
\begin{table}[t]
\centering
\footnotesize
\setlength{\tabcolsep}{6pt}
\renewcommand{\arraystretch}{1.15}
\caption{Starting-entity grounding accuracy on ReasonVQA. Strict accuracy requires the predicted entity QID to match the gold QID, while soft accuracy additionally accepts predictions whose entity label has SBERT similarity $\geq 0.7$ to the gold label.}
\label{tab:grounding}
\begin{tabular}{l|cc}
\toprule
\textbf{Topic-linking configuration} & \textbf{Strict} & \textbf{Soft} \\
\midrule
Top-1 Selection (Qwen2.5-VL-72B)         & 24.1 & 25.1 \\
\midrule
\methodname{} (Qwen2.5-VL-72B) & 42.7 & 48.3 \\
\methodname{} (GPT-4o)      & 56.2 & 63.5 \\
\methodname{} (GLM-4.6V)     & \textbf{57.2} & \textbf{64.8} \\
\bottomrule
\end{tabular}
\end{table}

%% file: tables/ablation.tex
\begin{table}[t]
\centering
\footnotesize
\setlength{\tabcolsep}{5pt}
\renewcommand{\arraystretch}{1.15}
\caption{Component ablation on \methodname{}-Bench (SBERT@0.3). Each row removes one component from \methodname{}.}
\label{tab:ablation}
\begin{tabular}{l|cccc}
\toprule
\textbf{Variant} & \textbf{2-hop} & \textbf{3-hop} & \textbf{4-hop} & \textbf{Avg} \\
\midrule
\rowcolor{blue!8}
\methodname{} & 74.0 & 62.4 & 57.0 & \textbf{64.5} \\
\midrule
\quad w/o Graph-aware validation             & 62.5 & 55.2 & 49.7 & 56.2 \\
\quad w/o Visual intent          & 70.1 & 56.3 & 51.6 & 59.0 \\
\quad w/o Reasoning-chain pruning       & 66.7 & 56.4 & 53.0 & 58.5 \\
\quad w/o Evidence checking & 74.0 & 61.6 & 58.0 & 64.2 \\
\bottomrule
\end{tabular}
\end{table}

%% file: tables/cost.tex
\begin{table}[t]
\centering
\footnotesize
\setlength{\tabcolsep}{4pt}
\renewcommand{\arraystretch}{1.15}
\caption{Per-question cost on \methodname{}-Bench: text-LLM calls, vision calls, text tokens, and wall-clock time.}
\label{tab:cost}
\begin{tabular}{l|cccc}
\toprule
\textbf{Method} & \textbf{LLM calls} & \textbf{VLM calls} & \textbf{Tokens} & \textbf{Time (s)} \\
\midrule
VLM zero-shot      & 0 & 1 & $0.8$k & $4.7$ \\
ToG                & $15$ & 0 & $8.0$k & $29.8$ \\
PoG                & $4.8$ & 0 & $4.1$k & $22.1$ \\
MindMap            & $3$ & 0 & $3.0$k & $11.9$ \\
OmniSearch         & $4$ & $2$ & $6.0$k & $20.8$ \\
\midrule
\methodname{} (GPT-4o-mini) & $5.8$ & $5.2$ & $3.6$k & $27.3$ \\
\methodname{} (GPT-4o)      & $6.5$ & $6.4$ & $4.8$k & $53.4$ \\
\bottomrule
\end{tabular}
\end{table}

%% file: fig/hyperparam_fig.tex
\begin{figure}[t]
  \centering
  \subfloat[Edge retention $M_1$]{\includegraphics[width=0.345\linewidth]{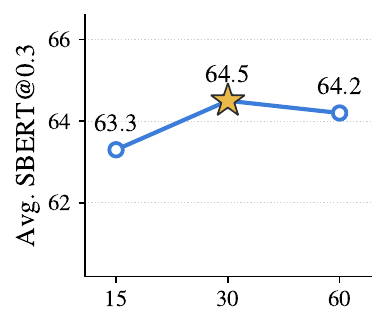}}
  \hfill
  \subfloat[Chain retention $M_3$]{\includegraphics[width=0.31\linewidth]{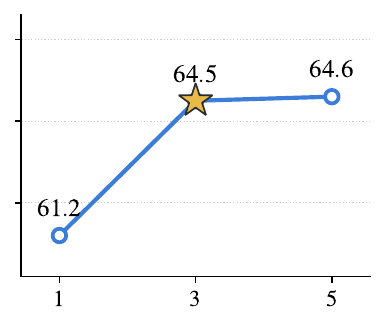}}
  \hfill
  \subfloat[Depth budget $D_{\max}$]{\includegraphics[width=0.31\linewidth]{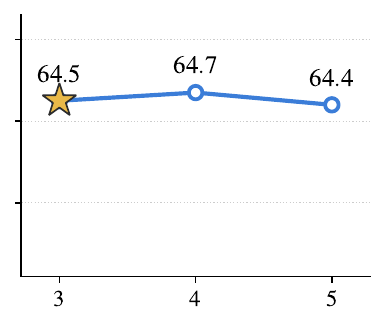}}
  \caption{Sensitivity to the main hyperparameters on \methodname{}-Bench
  (Avg.\ SBERT@0.3). Stars mark the default setting.}
  \label{fig:hyperparam}
\end{figure}

%% file: sec/6_conclusion.tex
\section{Conclusion}
In this paper, we focus on multimodal knowledge graph question answering (MM-KGQA), where a model must jointly ground multimodal input and perform multi-hop reasoning over structured KG evidence. We propose \methodname{}, a visual-intent-guided path reasoning framework. \methodname{} keeps multimodal information involved throughout the reasoning process by combining graph-aware starting entity identification, hop-specific intent-guided path discovery, reasoning-chain pruning, and evidence checking before answer generation. This design allows multimodal evidence to support not only entity grounding, but also relation selection, path filtering, and answer verification. We also introduce \methodname{}-Bench, a benchmark designed to evaluate multimodal multi-hop reasoning over KGs. Experiments on \methodname{}-Bench and three other multimodal QA benchmarks show that \methodname{} consistently improves over strong baselines. These results demonstrate the importance of keeping multimodal evidence in the reasoning loop for faithful MM-KGQA.

A promising future direction is to extend \methodname{} to temporal MMKGQA, where both multimodal evidence and KG facts evolve over time. In this setting, models need to identify not only visually grounded entities and reasoning paths, but also temporally valid evidence. Extending \methodname{} with time-aware multimodal intent may further support faithful reasoning over dynamic MMKG.